\documentclass[journal,twoside,web]{ieeecolor}
\usepackage{generic}
\usepackage{cite}
\usepackage{makecell}
\usepackage{amsmath,amssymb,amsfonts}
\usepackage{algorithmic}
\usepackage{tabularx}
\usepackage{booktabs}
\usepackage{graphicx}
\usepackage{algorithm}
\usepackage{hyperref}
\hypersetup{hidelinks}
\usepackage{textcomp}
\def\BibTeX{{\rm B\kern-.05em{\sc i\kern-.025em b}\kern-.08em
    T\kern-.1667em\lower.7ex\hbox{E}\kern-.125emX}}
\begin{document}
\title{Semi-Supervised Domain Adaptation with Latent Diffusion for Pathology Image Classification}

\author{Tengyue Zhang, Ruiwen Ding, Luoting Zhuang, Yuxiao Wu, Erika F. Rodriguez, William Hsu,~\IEEEmembership{Senior~Member,~IEEE} 
\thanks{This work was supported in part by NIH/National Cancer Institute U2C CA271898, U01 CA233370, and the V Foundation. (Corresponding author: William Hsu; e-mail: whsu@mednet.ucla.edu.)}
\thanks{T. Zhang, R. Ding, L. Zhuang, and W. Hsu are with Medical \& Imaging Informatics, Department of Radiological Sciences, David Geffen School of Medicine, University of California, Los Angeles (UCLA), Los Angeles, CA 90024, USA. (e-mail: olivia.zhang@ucla.edu; dingrw@ucla.edu; luotingzhuang@g.ucla.edu; whsu@mednet.ucla.edu). }
\thanks{Y. Wu is with the Bioengineering Department, Henry Samueli School of Engineering, UCLA, Los Angeles, CA 90024, USA (e-mail: yuxiaowu@mednet.ucla.edu).}
\thanks{E. F. Rodriguez is with the Department of Pathology \& Laboratory Sciences, David Geffen School of Medicine, UCLA, Los Angeles, CA 90024, USA (e-mail: erikarodriguez@mednet.ucla.edu).}}

\maketitle

\begin{abstract}
Deep learning models in computational pathology often fail to generalize across cohorts and institutions due to domain shift. Existing approaches either fail to leverage unlabeled data from the target domain or rely on image-to-image translation, which can distort tissue structures and compromise model accuracy. In this work, we propose a semi-supervised domain adaptation (SSDA) framework that utilizes a latent diffusion model trained on unlabeled data from both the source and target domains to generate morphology-preserving and target-aware synthetic images. By conditioning the diffusion model on foundation model features, cohort identity, and tissue preparation method, we preserve tissue structure in the source domain while introducing target-domain appearance characteristics. The target-aware synthetic images, combined with real, labeled images from the source cohort, are subsequently used to train a downstream classifier, which is then tested on the target cohort. The effectiveness of the proposed SSDA framework is demonstrated on the task of lung adenocarcinoma growth pattern classification. The proposed augmentation yielded substantially better performance on the held-out test set from the target cohort, without degrading source-cohort performance. The approach improved the weighted F1 score on the target-cohort held-out test set from 0.611 to 0.706 and the macro F1 score from 0.641 to 0.716. Our results demonstrate that target-aware diffusion-based synthetic data augmentation provides a promising and effective approach for improving domain generalization in computational pathology. Code is available at \url{https://github.com/hsu-lab/ldm-pathology-ssda}. 
\end{abstract}


\section{Introduction}
\label{sec:introduction}
\IEEEPARstart{D}{eep} learning (DL) models for computational pathology often exhibit limited generalizability across institutions due to domain shift arising from variations in staining protocols, scanner hardware, and tissue preparation method~\cite{sahiner2023data,jahanifar2025domain}. As a result, DL models that demonstrate strong performance during internal validation often have degraded performance when they are tested on external cohorts~\cite{jiao2022staining,van2021deep}. This limited generalizability hinders the clinical deployment of DL models in pathology. 

Traditionally, domain shift in computational pathology has been mitigated by either stain normalization or stain augmentation approaches. Classical stain normalization approaches (e.g., Macenko~\cite{macenko2009method}, Vahadane~\cite{vahadane2016structure}) match the source and target color distributions by estimating stain vectors and mapping the stain concentrations of the source image to those of a reference image. A major limitation of these methods is their reliance on a single reference image for normalization; the choice of reference can substantially affect the appearance of normalized images, resulting in poor reproducibility across cohorts. Stain augmentation avoids the reliance on a single image by applying random perturbations to the stain vectors during training to improve robustness~\cite{jahanifar2025domain}. However, these perturbations are not guided by information from the target domain, making these approaches suboptimal when such information is available. 

Addressing domain shift, therefore, requires incorporating target-domain information during training. This motivates adaptation approaches that can utilize unlabeled data from the target domain during training and do not alter target images at test time. Computational pathology is particularly challenging, as tile-level annotations require substantial time from pathologists and are often subject to inter-reader variability~\cite{ding2023tailoring}. Consequently, in real-world settings, target-domain data often do not provide expert-annotated labels. Formally, this problem aligns with the semi-supervised domain adaptation (SSDA) problem, where we have: (1) labeled data and unlabeled data from the source domain, (2) unlabeled data from the target domain, and (3) no target images modified at test time. 

Existing SSDA methods in computational pathology fall into two major families. The first uses Generative Adversarial Networks (GAN)-based style-transfer methods to map the labeled source images to the style of the target images~\cite{cong2022colour,jose2021generative,kang2021stainnet,xu2019gan}. Yet, they can distort or hallucinate tissue morphology, thereby degrading model accuracy~\cite{du2025deeply,pezoulas2024synthetic}. The second type of approach focuses on learning stain-invariant~\cite{otalora2019staining,marini2021h} or domain-invariant~\cite{lafarge2019learning,fang2023domain} features via adversarial training. These methods minimize the distance between source and target feature distributions globally, without distinguishing between variations due to domain shift and variations due to biological differences. In tasks that require the model to have discriminative ability of subtle morphological differences, this global alignment might reduce discriminative features, thereby having suboptimal performance~\cite{AIDA, xiao2021simultaneously}. Furthermore, these methods have been predominantly validated on large, labeled source datasets that contain hundreds of WSIs~\cite{AIDA,lafarge2019learning,fang2023domain}. Annotations for this many WSIs would require substantial time from pathologists and may not be available in many clinical settings. 

Diffusion models have demonstrated a strong ability to generate realistic pathology images~\cite{yellapragada2025zoomldm,graikos2024learned,yellapragada2024pathldm,alfasly2025semantic}. They can generate synthetic images with variations from the real source images while preserving the morphology and semantics of the source images, making them suitable for data augmentation. However, these works primarily evaluate the quality of the diffusion-generated synthetic images or their ability to perform in-domain augmentation. More recently, diffusion models have been applied to the domain adaptation task. Tsai et al.~\cite{tsai2024test} proposed using a diffusion model to perform stain adaptation at test time. However, as noted above, such test-time adaptation can distort tissue structure and compromise model performance. MeDi~\cite{drexlin2025medi} conditions a diffusion model on class labels and metadata such as hospital site to augment underrepresented subgroups during training, but it requires labels for all training images, including those from both source and target domains. No prior study uses a diffusion model trained on unlabeled data to generate synthetic images for cross-cohort generalization.

In this work, we propose an SSDA framework in which a latent diffusion model (LDM) is trained on unlabeled data from both the source and target domains and used to generate synthetic images, thereby improving the downstream model's generalization to the target domain. Leveraging the ability of diffusion models to generate diverse images, we create structural variations while preserving the underlying tissue morphology. At the same time, by training the LDM on both domains, we shape a target-aware diffusion prior that captures target-domain appearance (e.g., staining, color, scanner-related differences, tissue preparation-related differences such as tissue thickness), allowing the synthetic images to resemble the target-domain appearance. The LDM is conditioned on feature embeddings extracted from the real source images to preserve tissue morphology, enabling the generation of labeled synthetic images using an LDM trained on unlabeled data. The LDM-generated synthetic images are combined with labeled images from the source domain to train a downstream classifier. We hypothesize that the downstream classifier trained on real source-domain images augmented with the target-aware synthetic images will have better generalization to the target domain, outperforming conventional augmentation methods. 

Our contributions are as follows: 
\begin{itemize}
    \item We introduce a novel diffusion-based SSDA framework that leverages unlabeled source- and target-domain images to generate target-aware synthetic data.
    \item We design a pathology-specific conditioning mechanism that incorporates image embeddings, cohort, and tissue preparation to generate morphology-preserving and target-aware synthetic images. 
    \item We demonstrate the effectiveness of the LDM-based augmentation approach on the lung adenocarcinoma growth pattern classification task, showing improvements in the generalization of the downstream model while maintaining the performance of the source cohort. 
\end{itemize}

\section{Methods}

\subsection{Overview}

\subsubsection{Problem formulation}In the SSDA problem formulation, we have: (1) $\mathcal{D}_s^\text{unlabeled} = \{x_j\}$, the unlabeled data from the source domain; (2) $\mathcal{D}_s^\text{labeled} = \{(x_i, y_i)\}$, the labeled data from the source domain; (3) $\mathcal{D}_t^\text{unlabeled} = \{x_k\}$, the unlabeled data in the target domain. The objective is to use data from the source domain (i.e., $\mathcal{D}_s^\text{unlabeled}$ and $\mathcal{D}_s^\text{labeled}$) and information about the target domain provided by $\mathcal{D}_t^\text{unlabeled}$ to build a classifier that generalizes well to the target domain. 

\subsubsection{Framework overview}
\begin{figure*}[!t]
\centerline{\includegraphics[width=18.5cm]{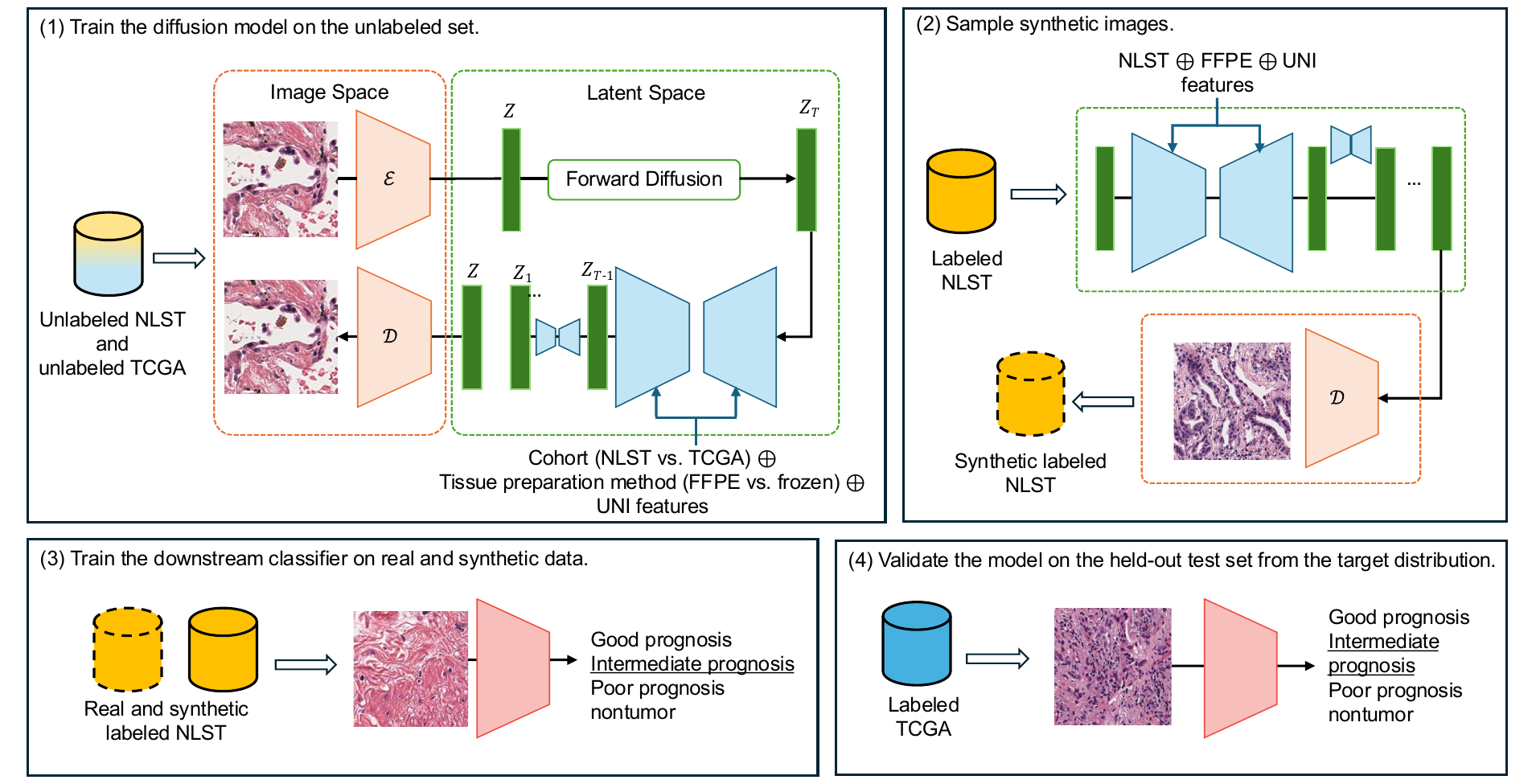}}
\caption{Overview of the proposed synthetic-augmentation semi-supervised domain adaptation pipeline. The latent diffusion model (LDM) is trained on unlabeled source-domain images (e.g., NLST) and unlabeled target-domain images (e.g., TCGA). Synthetic images are generated from the UNI features of the real, labeled images from NLST. These synthetic NLST tiles are combined with real labeled NLST tiles to train a downstream classifier. The downstream classifier is evaluated on a held-out set of labeled TCGA images, which does not contain patients used for LDM training.}
\label{fig:overview}
\end{figure*}

The overview of our approach is shown in Fig.~\ref{fig:overview}, consisting of three steps. First, an LDM is trained on $\mathcal{D}_s^\text{unlabeled}$ and $\mathcal{D}_t^\text{unlabeled}$, letting the model learn a generative prior that captures the tissue appearance in the target domain. Second, the LDM is used to generate synthetic source images in a target-aware manner. For each image in $\mathcal{D}_s^\text{labeled}$, a synthetic counterpart is generated by conditioning the LDM on a feature embedding extracted from the real source image. Since the LDM is trained on unlabeled images from both domains, the synthetic counterparts may include the appearance characteristics of the target domain. Third, a classifier is trained on real labeled source images combined with target-aware synthetic images. Because the synthetic images include the image appearance of the target domain, the classifier becomes more generalizable to the target domain. 

\subsection{Latent diffusion model}

\subsubsection{LDM architecture}The LDM consists of a variational autoencoder (VAE), a U-Net denoiser, and a conditioning branch~\cite{rombach2022high}. The VAE encoder converts an input image to a latent space, and the VAE decoder reconstructs the image from the latent representations. The U-Net denoiser performs the reverse diffusion process in the latent space. It predicts the noise added to the latents given a diffusion timestep and the conditions. This setup enables a more efficient diffusion process in the latent space compared to the image space. 

In the conditioning branch, we condition the model on three types of information to preserve tissue morphology and incorporate domain-specific variations. For each tile, we obtain: (1) a feature embedding extracted from UNI, a pathology image foundation model~\cite{chen2024towards}; (2) cohort; and (3) tissue-preparation method, i.e., formalin-fixed paraffin-embedded (FFPE) or frozen tissue. The cohort and tissue preparation conditions are one-hot encoded (e.g., 0 = FFPE, 1 = frozen) and then converted into continuous values through embedding layers. Each of these three condition embeddings is passed through a linear layer and concatenated to form the final condition vector. The condition vector is injected into the U-Net via cross-attention layers. The latent representations from the U-Net serve as queries, and the condition vector provides keys and values, enabling the image features to learn to attend to the information in the condition vector. Through this cross-attention mechanism, the U-Net denoiser is guided by image-level and cohort-level contexts provided by the conditioning information. We selected this concatenation-and-cross-attention conditioning mechanism based on an ablation study comparing three alternative designs: concatenation-based cross-attention, cascaded cross-attention, and multi-token cross-attention (Supplementary Table S1). 

\subsubsection{Training the LDM}The LDM is trained on unlabeled source and target images, i.e., $\mathcal{D}_s^\text{unlabeled} \cup \mathcal{D}_t^\text{unlabeled}$. We follow the standard denoising diffusion probabilistic model training setup~\cite{rombach2022high}. For a training image $x$, we obtain the latent representation $z$ from the VAE encoder and perturb $z$ with Gaussian noise to obtain $z_t$, the noisy latent at diffusion timestep $t$. The U-Net denoiser is trained to predict the injected noise $\epsilon$ and is optimized by the diffusion loss: 
\begin{equation}
   L_\text{diff} = \mathbb{E}_{z, \epsilon, t} \big[ || \epsilon - \epsilon_\theta(z_t, t, c) || ^2  \big]
\end{equation}
where $c$ is the conditions, $t$ is the diffusion timestep, and $\epsilon_\theta(z_t, t, c)$ is the U-Net denoiser's prediction. 

\subsection{Generate synthetic images from the LDM}

To generate synthetic images useful for downstream classification, we create a synthetic counterpart for each real, labeled image in the source domain. For each image in $\mathcal{D}_s^\text{labeled}$, UNI features were extracted and used to condition the LDM. Because the UNI features used for conditioning encode morphology in the source image, we assume that the synthetic counterpart preserves the same LUAD growth pattern as the original image. In this way, we can generate synthetic labeled images from an LDM trained on unlabeled images. Synthetic images are generated via the reverse diffusion process in the latent space, and the latent representation is decoded back into the image space using the VAE decoder. 

\subsection{Combining synthetic images with stain augmentation}\label{sec:combined_aug}

We further augment the synthetic images generated by the LDM using traditional stain augmentation approaches, namely Macenko stain augmentation~\cite{TELLEZ2019101544,macenko2009method} and Vahadane stain augmentation~\cite{TELLEZ2019101544,vahadane2016structure}. These stain augmentation approaches work by performing stain separation, applying perturbations to the stain vectors, and reconstructing the image from the perturbed stain vectors. Macenko and Vahadane augmentations differ in the way they perform stain separation. Macenko uses Singular Value Decomposition to convert an image into an optical density space, whereas Vahadane uses sparse non-negative matrix factorization to extract stain components. After generating the synthetic images, we apply these stain augmentation methods to both real and synthetic images when training the downstream classifier. We use this combined strategy to expose the downstream model to the image appearance and tissue structure learned by the LDM while also introducing more aggressive variability in color and staining through traditional stain augmentation.

\subsection{Downstream classification task \& model}

\subsubsection{Downstream classification task}
We investigate the SSDA problem through the lung adenocarcinoma (LUAD) histologic growth pattern classification task, where each pathology image is classified as `good prognosis', `intermediate prognosis', `poor prognosis', or `nontumor', based on the histologic growth patterns defined by the International Association for the Study of Lung Cancer (IASLC) pathology committee~\cite{moreira2020grading}. The class labels are named after the prognostic outcomes of each growth pattern. The model predicts the growth pattern of each tile, not patient-level outcomes. The `good prognosis' class contains tumors with the lepidic growth pattern. Patients with lepidic-predominant tumors have a 5-year survival of 93\%. The `intermediate prognosis' class contains tumors with acinar and papillary patterns. Patients with acinar- and papillary-predominant patterns have 5-year survival of 71\% and 68\%, respectively. Solid and micropapillary patterns are considered to have a poor prognosis, with 5-year survival rates of 39\% and 38\%~\cite{russell2011does}. Classifying into these categories is clinically significant as patients with a worse prognosis are more likely to benefit from adjuvant chemotherapy~\cite{moreira2020grading}. Similar to other pathology tasks, LUAD growth pattern classification is also susceptible to domain shifts~\cite{al2025cellomaps,ding2023tailoring}. 

\subsubsection{Downstream classification model}
The downstream classification model is trained on the real labeled tiles $\mathcal{D}_s^\text{labeled}$ augmented by the LDM-generated target-aware synthetic tiles. For each synthetic tile, we assign the same label as its corresponding source tile. The classifier is trained by fine-tuning the Vision Transformer (ViT) architecture~\cite{dosovitskiy2020image}. The ViT outputs a probability corresponding to each LUAD growth pattern and classifies an image into the class with the highest probability. A cross-entropy loss is used to supervise training. When using the synthetic images, we balance the influence of synthetic data on model training by weighting the contributions of the loss terms computed on real and synthetic images. The total loss is:
\begin{equation}
    L_\text{downstream} = w_s\cdot L_\text{CE}^s + (1-w_s) \cdot L_\text{CE}^r
\end{equation}
where $L_\text{CE}^s$ and $L_\text{CE}^r$ denote the cross-entropy loss computed on synthetic and real images, respectively. $w_s$ is found via a grid search. 

\section{Experiments}

\subsection{Data}

\subsubsection{Datasets}

\begin{table}[t]
\centering
\renewcommand{\arraystretch}{1.3}
\begin{tabular}{lp{1.3cm}p{0.9cm}p{1.2cm}p{1.2cm}p{0.6cm}}
\hline
\textbf{Cohort} & \textbf{\# patients} & \textbf{\# WSIs} &
\textbf{\# females} & \textbf{\# stage I} & \textbf{Mean age} \\
\hline
NLST & 190 & 536 & 89 (46.8\%) & 155 (81.6\%) & 63.7 \\
TCGA & 399 & 1023 & 216 (54.1\%) & 275 (68.9\%) & 65.8 \\
\hline
\end{tabular}
\caption{Summary statistics for patient characteristics.}
\label{tab:datasets}
\end{table}

Hematoxylin and Eosin (H\&E)-stained whole-slide images (WSIs) were obtained from two publicly available datasets: NLST~\cite{nlst} (190 patients, 536 WSIs) and TCGA~\cite{tcga} (399 patients, 1,023 WSIs). For both datasets, patients were selected such that they had Stage I or Stage II LUAD and at least one WSI available. Characteristics of the patient cohorts are shown in Table~\ref{tab:datasets}. In our experiments, NLST is treated as the source domain, and TCGA is treated as the target domain. 

\subsubsection{Data preprocessing}
WSIs are tiled into patches using TRIDENT~\cite{vaidya2025molecular,zhang2025standardizing}. Tissue segmentation and penmark removal were first performed on WSIs, and the remaining tissue was tiled into non-overlapping patches. For WSIs scanned at 40$\times$ objective magnification, tiles were extracted at 10$\times$ magnification with size $512 \times 512$. For WSIs scanned at 20$\times$ objective magnification, tiles were extracted at 20$\times$ magnification with size $1024 \times 1024$ so that the physical area covered by each tile is consistent across WSIs with different objective magnification. 

\subsubsection{Annotation}\label{sec:annotation}
Tile-level annotations were originally conducted by two pathologists, following the procedure described in~\cite{ding2023tailoring}. Briefly, each WSI was presented as a grid of tiles. The pathologists had access to the full multi-resolution WSIs and selected representative tiles of each growth pattern. When one pathologist was uncertain of a label, the tile was independently reviewed by the second pathologist for consensus. Inter-reader agreement was analyzed on 30 tiles requiring consensus, yielding a Cohen’s kappa of 0.72 (95\% CI: 0.53-0.92). In this work, each annotated tile was further divided into four non-overlapping sub-tiles. To verify that labels accurately carried over to the sub-tiles, a random sample of 15 sub-tiles was reviewed by E.R., and all of them were confirmed to have the same labels as the original annotated tiles. This consistency is expected as, in the original annotation procedure, only tiles containing a single predominant growth pattern were annotated, resulting in sub-tiles that contain the same morphological patterns of the assigned class. 

\subsubsection{Data splits}The number of tiles for LDM and downstream classification is shown in Table~\ref{tab:data_splits}. To prevent data leakage, there is no overlap in patient cases with unlabeled tiles used for LDM development and the cases with labeled tiles used for downstream model development or testing. This setup simulates the SSDA setting where domain adaptation is performed on unlabeled data from the target domain, and the downstream classifier is tested on previously unseen patients.

Patients with unlabeled data from NLST and TCGA were split into 80\% training and 20\% validation, with no overlapping patient cases between the two sets. Tiles from 80\% of patients are used for training the LDM, and the fidelity of the synthetic images is compared against the tiles from the 20\% validation patients. 

For the downstream classification task, we used 2,638 labeled tiles from NLST for cross-validation and 871 labeled tiles from TCGA for testing. To provide a robust evaluation of model performance, we used 5-fold cross-validation with a patient-level split. In each fold, 20\% of patients from NLST were randomly selected as the test set. A multilabel stratified splitting was used to approximately preserve the class distribution across folds. The tiles from the remaining 80\% of NLST patients were split into 80\% training and 20\% validation at the image level. Each model is trained and evaluated across all five folds, and performance is reported on both the cross-validation test sets (NLST) and the held-out test set from TCGA.

\begin{table*}[ht]
\centering
\renewcommand{\arraystretch}{1.3}
\begin{tabular}{l c c c c c c}
\hline
 & \multicolumn{2}{c}{\textbf{Diffusion (unlabeled)}} 
 &  & \multicolumn{3}{c}{\textbf{Downstream (labeled)}} \\[4pt]
\cline{2-3} \cline{5-7}
\textbf{Cohort} 
& Train & Validation
&  & Train & Validation & Test \\
\hline
NLST  
& 1,256,490 & 311,355 
&  & 1716 & 429 & 493 \\
TCGA  
& 1,215,329 & 275,820 
&  & 0    & 0   & 871 \\
\hline
Total 
& 2,471,819 & 587,175 
&  & 1716 & 429 & 
\begin{tabular}[c]{@{}l@{}}493 from the CV test set \\ 871 from the held-out test set\end{tabular} \\
\hline
\end{tabular}
\caption{Number of tiles used for diffusion model training/validation and downstream classifier training/validation/testing. There are no overlapping patient cases between the unlabeled set for diffusion development and the labeled set for downstream to simulate the semi-supervised domain adaptation setting. }
\label{tab:data_splits}
\end{table*}

\begin{table*}[h!]
\centering
\renewcommand{\arraystretch}{1.3}
\begin{tabular}{lccccc}
\hline
 & Good prognosis & Intermediate prognosis & Poor prognosis & Non-tumor & Total \\
\hline
NLST & 456 (17.29\%) & 910 (34.50\%) & 720 (27.29\%) & 552 (20.92\%) & 2{,}638 (100\%) \\
TCGA & 114 (13.09\%) & 271 (31.11\%) & 271 (31.11\%) & 215 (24.68\%) & 871 (100\%) \\
\hline
\end{tabular}
\caption{Number of tiles of each lung adenocarcinoma prognostic category in the downstream labeled set.}
\label{tab:subtypes}
\end{table*}

\subsection{Baselines and ablation studies}
We evaluate our approach by comparing it with three types of baselines: (1) image and stain augmentation methods, (2) foundation model features, and (3) representative methods from other families of SSDA approaches, namely CycleGAN-based stain normalization and feature alignment methods. All models are evaluated by the same NLST cross-validation test sets and TCGA held-out test set. 

The models compared are: 
\begin{enumerate}
    \item \textbf{No augmentation}: The model is trained on the training images without applying any data augmentation. 
    \item \textbf{Geometric transformations}: Geometric transformations, including random flips and rotations, were applied during training.
    \item \textbf{Color jittering}: Perturbations in brightness, contrast, hues, and saturation were applied to the training images to simulate variations in color. 
    \item \textbf{Standard augmentation}: Both geometric transformations and color jittering were applied. 
    \item \textbf{Macenko stain augmentation}~\cite{TELLEZ2019101544,macenko2009method}: Macenko stain augmentation first decomposes the H\&E-stained image into H stain and E stain, applies perturbations to the stain concentrations, and reconstructs the image. 
    \item \textbf{Vahadane stain augmentation}~\cite{TELLEZ2019101544,vahadane2016structure}: Vahadane augmentation follows the same process as Macenko, except that it uses sparse non-negative matrix factorization to separate the H and E stains, which results in more accurate stain separation. 
    \item \textbf{UNI + logistic regression}: In addition to the data augmentation baselines, we also include a foundation-model baseline to assess whether representations learned by pre-trained pathology foundation models can mitigate cross-cohort domain shifts. In this model, tile-level features are extracted from UNI~\cite{chen2024towards}. Following the linear probing approach suggested by the authors, we fit a multinomial logistic regression model on NLST features and evaluate it on the held-out test set from TCGA. 
    \item \textbf{CycleGAN stain normalization}~\cite{runz2021normalization}: CycleGAN stain normalization learns an unpaired image-to-image mapping between the source domain (NLST) and target domain (TCGA) using a cycle-consistent generative adversarial network. Following the original work~\cite{runz2021normalization}, we re-trained the CycleGAN on our unlabeled data from source and target domains and augmented training images with their CycleGAN-normalized versions. 
    \item \textbf{Adversarial Fourier-based domain adaptation (AIDA)}~\cite{AIDA}: AIDA applies adversarial training in the Fourier frequency domain to align the feature distributions of the source and target domains, achieving state-of-the-art performance on cancer classification tasks. As with our method, AIDA addresses the domain adaptation problem under the same setting: when labeled source data and unlabeled target data are available for training, with all labeled target data being held out for testing. Therefore, we used AIDA as a representative method from the feature alignment family. 
    \item \textbf{LDM synthetic augmentation (proposed)}: This model uses synthetic images generated by the LDM to augment the training data. 
    \item \textbf{LDM synthetic + Macenko (proposed)}: This model applies Macenko stain augmentation to the real and LDM-generated synthetic images during training as described in Sec.~\ref{sec:combined_aug}.
    \item \textbf{LDM synthetic + Vahadane (proposed)}: This model applies Vahadane stain augmentation to the real and LDM-generated synthetic images during training as described in Sec.~\ref{sec:combined_aug}. 
\end{enumerate}

In addition to the baselines, we also conducted ablation studies to examine the effect of LDM training data on the generalization of the downstream model. Specifically, we compared models trained using synthetic images generated by an LDM trained on NLST+TCGA with those trained on NLST only. 

\subsection{Evaluation}

\subsubsection{Evaluating the LDM}To evaluate the quality and fidelity of LDM-generated synthetic images, the Fréchet Inception Distance (FID) of the Inception-V3 features~\cite{heusel2017gans} was computed on the validation split of unlabeled data. FID measures the similarity of the distributions of real and synthetic images. It assumes each set of features follows a multidimensional Gaussian distribution and computes the Fréchet distance between two sets of features. To calculate FID, we randomly selected 10,000 images from the validation split of the unlabeled data, maintaining the same proportion of NLST and TCGA cases as in the full validation split. For each selected image, UNI features, cohort, and tissue preparation information were extracted and used to generate a synthetic counterpart using the LDM. Finally, FID was calculated on the 10,000 real unlabeled validation images and the 10,000 condition-matched synthetic images. In addition, we computed the cosine similarity between real and synthetic images using features extracted from UNI~\cite{chen2024towards} and CONCH~\cite{lu2024visual} to provide a more pathology-specific assessment. Cosine similarity was computed on the same 10,000 image pairs used for FID evaluation.

\subsubsection{Evaluating downstream classifiers} The F1 score was used to measure the performance of downstream models. We used the F1 score because LUAD growth pattern classification is class-imbalanced, making accuracy a misleading metric. Moreover, the F1 score balances false positives and false negatives, consistent with the clinical setting, where both have critical consequences. In addition to per-class F1 scores, macro F1 (i.e., unweighted average) and weighted F1 (i.e., weighted by number of instances in each class) were calculated to provide an overall measure of model performance. 5-fold cross-validation F1 scores were compared between LDM+Vahadane and each baseline using a one-sided paired \textit{t}-test. To control for the false discovery rate across the eight baseline comparisons within each metric, p-values were adjusted using the Benjamini-Hochberg procedure at FDR = 0.05~\cite{benjamini1995controlling}. 

\subsubsection{Evaluating label preservation in synthetic images}\label{sec:label_preservation_method}In the proposed framework, one core assumption is that synthetic counterparts inherit the same LUAD growth pattern as their corresponding real source images because the UNI features encode tissue morphology. To verify this assumption, we conducted two analyses. In the first analysis, we first trained a logistic regression classifier on CONCH~\cite{lu2024visual} features from real images and applied the classifier to synthetic images. Then we measured the agreement between the classifier-predicted labels and the labels assigned to synthetic images based on their real source counterparts using Cohen's kappa. This evaluation was performed under the same 5-fold cross-validation splits as our downstream classification experiments to ensure consistency. Additionally, we computed the cosine similarity between CONCH features of real images and their synthetic counterparts within each class. CONCH features were used rather than UNI features, which were used for LDM conditioning, to ensure a more independent evaluation.

\subsection{Implementation details}

\subsubsection{LDM}We implemented the LDM using the PathLDM framework~\cite{yellapragada2024pathldm} and initialized the pre-trained VAE of PathLDM. The VAE is a Vector-Quantized Variational Autoencoder (VQ-VAE) with three latent channels and a downsampling factor of 4. The diffusion model follows a standard latent diffusion formulation, with 1,000 diffusion steps and a linear noise schedule. The U-Net denoiser was trained from scratch for 150,000 iterations, optimized using AdamW with a base learning rate of $5 \times 10^{-5}$ and cosine decay. Images were resized to $256\times 256$ pixels. An effective batch size of 48 was used, with 24 samples per GPU and 2 GPUs. In the classifier-free guidance (CFG), the conditioning information was dropped with a probability of 0.1 during training, allowing the model to learn both conditional and unconditional generation and enabling control over the tradeoff between image diversity and fidelity during inference. During inference, the CFG scale was set to 1.5. We conducted a sensitivity analysis on the cohort CFG scale ($\text{CFG}_\text{cohort}$), covering $\text{CFG}_\text{cohort} \in \{0.5, 1.0, 1.5, 3.0, 5.0 \}$ while holding the UNI CFG scale fixed. Downstream classification performance peaked at $\text{CFG}_\text{cohort} =1.5$, which is used for all the reported experiments. The results of the sensitivity analysis are in Supplementary Table S2. All training and inference were performed on NVIDIA L40S GPUs. 

\subsubsection{Downstream classifiers}For the downstream classification task, we fine-tuned an ImageNet~\cite{deng2009imagenet} pre-trained ViT-B/16 model~\cite{dosovitskiy2020image} from Hugging Face~\cite{wolf2019huggingface}. Images were resized to $224 \times 224$ and normalized by the mean and standard deviation of ImageNet statistics. Models were trained with the Adam optimizer~\cite{kingma2014adam}, learning rate $1 \times 10^{-4}$, weight decay $1 \times 10^{-4}$, and batch size 32. All models were trained for a maximum of 200 epochs, with an early stopping patience of 5, monitored by the validation loss. The hyperparameter $w_s$ was found via a grid search. We explored the range $w_s \in \{ 0.1, 0.25 , 0.5, 0.75, 1.0\}$ and selected $w_s=0.25$, which yielded the best validation performance. All downstream experiments were done on a single NVIDIA L40S GPU. 

\section{Results}

\begin{figure*}[ht]
\centerline{\includegraphics[width=18.5cm]{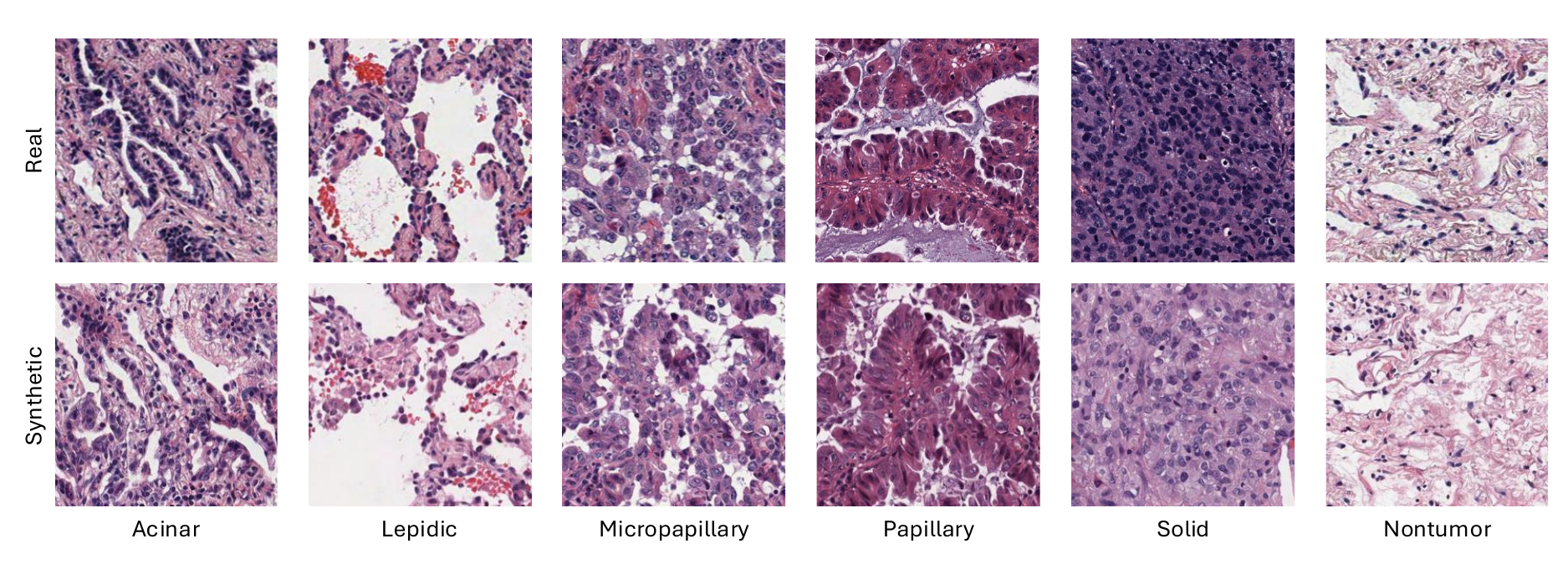}}
\caption{Examples of real and synthetic NLST images. For each real NLST image in the top row, a corresponding synthetic tile (bottom row) was generated from the LDM. The synthetic images preserve morphology while introducing variations in image appearance and tissue structures. Additional examples are provided in Supplementary Fig. S1.}
\label{fig:examples}
\end{figure*}

\subsection{LDM-generated synthetic images}

We evaluated the LDM using 10,000 real validation images and 10,000 synthetic images. The LDM achieved an FID of 5.779. In practice, FID values below 10 are generally considered indicative of high fidelity. Our FID value reveals a strong similarity between the real and synthetic image distributions, suggesting that the LDM generates high-fidelity images that accurately capture the characteristics of LUAD pathology images. Additionally, the UNI cosine similarity was 0.748, and the CONCH cosine similarity was 0.861, indicating strong feature-level similarity between real and synthetic images. Representative examples are shown in Fig.~\ref{fig:examples}.

\subsection{Label preservation in synthetic images}\label{sec:label_preservation}

As described in Section~\ref{sec:label_preservation_method}, we validated the label preservation through an independent classifier and a cosine similarity analysis. The logistic classifier trained on CONCH features achieved a Cohen’s kappa of $0.744 \pm 0.062$. Notably, this value is comparable to the inter-reader agreement ($\kappa =0.72$) between two pathologists on the 30 most ambiguous cases in our annotated dataset. This finding suggests that synthetic images preserve subtype-discriminative features with a reliability level similar to human expert annotation. In the cosine similarity analysis, as shown in Table~\ref{tab:conch_label_preservation}, cosine similarities were around or above 0.8 in all four classes. This is higher than the cosine similarity values computed between two randomly sampled images within the same class and two randomly sampled images of two different classes (Supplementary Table S3). This finding indicates that synthetic images remain closely aligned with their real source images in the feature space. Together, these two analyses support the label preservation assumption and show that synthetic images generated by the LDM retain the subtype-discriminative morphology of their source images. 

\begin{table}[h!]
\begin{minipage}{\columnwidth}
\centering
\renewcommand{\arraystretch}{1.3}
\begin{tabular}{lc}
\hline
\textbf{Class} & \textbf{Cosine Similarity (mean $\pm$ SD)} \\
\hline
Good prognosis & $0.805 \pm 0.110$ \\
Intermediate prognosis	& $0.816 \pm 0.074$ \\
Poor prognosis & $0.790 \pm 0.088$ \\
Non-tumor &	$0.812 \pm 0.094$ \\
\hline
\end{tabular}
\caption{Mean cosine similarity of CONCH feature embeddings between real source images and their synthetic counterparts, computed within each LUAD growth pattern. SD: standard deviation.} 
\label{tab:conch_label_preservation}
\end{minipage}
\end{table}

\begin{figure*}[ht]
\centerline{\includegraphics[width=18.5cm]{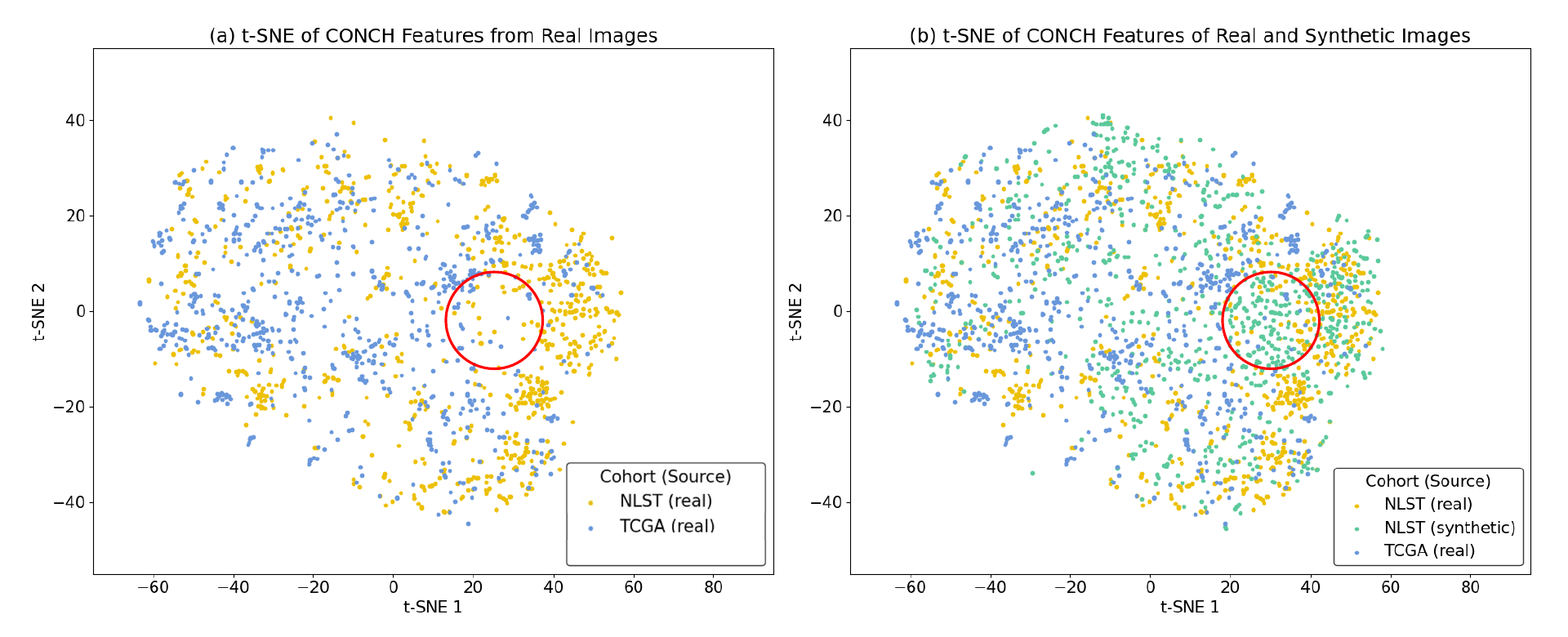}}
\caption{t-SNE visualization of real NLST, real TCGA, and synthetic NLST tiles. Plots are generated using CONCH feature embeddings. (a) shows the t-SNE feature space of real NLST and real TCGA tiles. (b) shows the t-SNE with real NLST, real TCGA, and synthetic NLST tiles. While real NLST and real TCGA images do not exhibit completely separated clusters, there still exist gaps between the two domains, as highlighted by the red circle. Synthetic NLST tiles generated by the LDM occupy an intermediate space between NLST and TCGA distributions, indicating that using unlabeled TCGA during diffusion training influences the generative prior and introduces target-domain characteristics.}
\label{fig:t_sne}
\end{figure*}

\subsection{Synthetic images enable target-aware augmentation}

To assess whether the synthetic images generated by the LDM can capture the image appearance of the target cohort, we performed t-SNE projections of embeddings from CONCH, a vision-language histopathology foundation model~\cite{lu2024visual}, to compare the feature distribution of real and synthetic images. Since UNI embeddings were used for conditioning, t-SNE projections of the UNI embeddings are expected and, in fact, show substantial overlap between real and synthetic images. This overlap shows the LDM preserves information in the UNI features but is not informative for assessing the additional variations provided by the LDM. Therefore, we used CONCH embeddings to visualize the domain gap bridged by the LDM-generated synthetic data. 800 unlabeled NLST and 800 unlabeled TCGA tiles were selected from the validation set, and synthetic counterparts of the 800 NLST images were generated using the LDM. As shown in Fig.~\ref{fig:t_sne}(a), while the two cohorts do not exhibit completely separated clusters, there is a noticeable gap between the feature space of NLST images and that of TCGA images in certain areas, as indicated by the red circle. Fig.~\ref{fig:t_sne}(b) shows the t-SNE projection after adding synthetic NLST images. The synthetic images fill the gaps between the original NLST and TCGA feature spaces, interpolating the two distributions. This indicates that the synthetic tiles capture characteristics more similar to the TCGA cohort compared to the original real NLST tiles.

\subsection{Downstream classification performance}\label{sec:downstream_results}

\subsubsection{Held-out test set from the target cohort}

To assess whether the downstream classifiers generalize to the target cohort, we evaluated all models on the held-out labeled TCGA test set, which consists entirely of patients unseen in LDM training and evaluation. All three variants of the LDM-based augmentation models improved the generalization to TCGA, indicating the effectiveness of diffusion-based data augmentation. The model using LDM+Vahadane augmentations achieved the best overall F1 scores (weighted 0.706, macro 0.716) and the best F1 scores on the `intermediate prognosis' (0.710) and `poor prognosis' (0.592) classes. LDM synthetic augmentation achieved the best F1 scores in the `good prognosis' (0.769) and `nontumor' (0.835) classes while achieving the second-best macro F1 (0.701). 

In the baseline models, geometric transformation and Macenko stain augmentation yielded moderate improvements in model performance on the TCGA held-out test set, resulting in higher overall F1 scores compared to the model with no augmentation. Among all baseline models, Vahadane augmentation showed the largest improvement over the no-augmentation model, achieving a weighted F1 score of 0.647; however, the difference was not statistically significant ($p > 0.5$). Color jittering and standard augmentation worsened model performance on TCGA, resulting in lower overall F1 scores compared to the no-augmentation model. Notably, the logistic regression model using foundation model features did not outperform fine-tuned ViT when stain and/or LDM-based augmentation were used. This indicates that, although UNI offers robust general-purpose representations, linear probing of UNI frozen features is insufficient to overcome cross-domain and cross-cohort differences. Among the CycleGAN and AIDA baselines, CycleGAN achieved competitive performance on the target-cohort test set (weighted: 0.654, macro: 0.675). This finding is consistent with prior findings: CycleGAN-based image translation methods can distort tissue morphology, which may limit their utility despite competitive target-domain performance~\cite{du2025deeply,pezoulas2024synthetic}. AIDA achieves competitive performance on both source- and target-cohort test sets (NLST source weighted: 0.737, macro: 0.700; TCGA target weighted: 0.634, macro: 0.651). However, the performance on the target cohort is suboptimal compared to stain augmentation and CycleGAN approaches. 

\begin{figure*}[htb]
\centerline{\includegraphics[width=18.5cm]{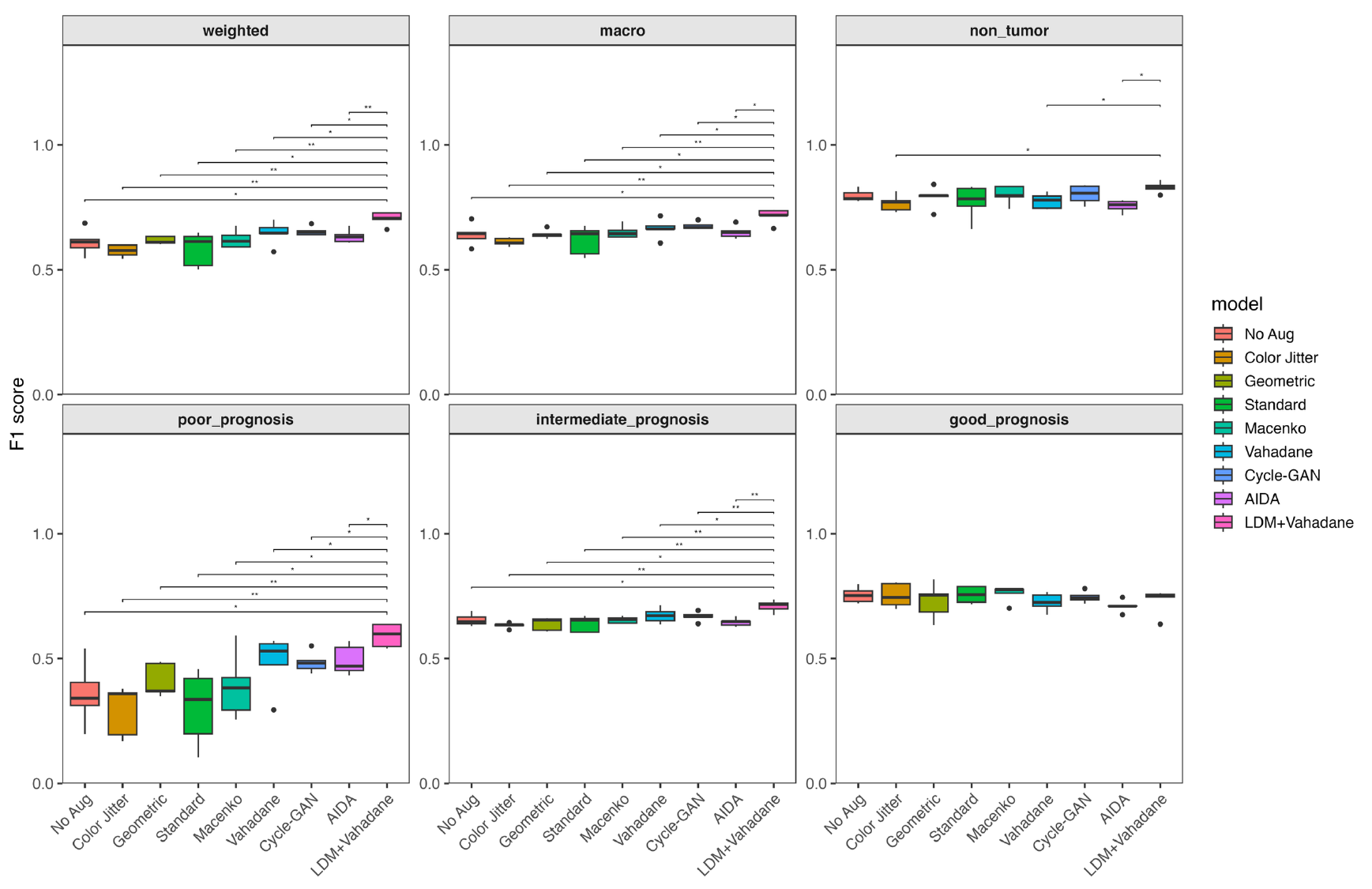}}
\caption{Box plots showing the distribution of F1 scores from 5-fold cross-validation when models using different data augmentation, normalization, and feature alignment approaches are tested on the TCGA held-out test set. Significance brackets indicate pairwise comparison between the proposed LDM+Vahadane method and each baseline via a one-sided paired t-test, with p-values adjusted for multiple comparisons using the Benjamini-Hochberg method (FDR = 0.05, m = 8 comparisons per metric). *$p_\text{adjusted}<0.05$, **$p_\text{adjusted} < 0.01$. Only significant comparisons are shown.}
\label{fig:box_plots}
\end{figure*}

Comparing the proposed LDM-based approaches with baseline methods, all three LDM-based models substantially improved the generalization to TCGA compared to the no-augmentation model. As shown in Fig.~\ref{fig:box_plots}, the LDM+Vahadane model yielded significantly better F1 scores than all baseline models in the `poor prognosis' and `intermediate prognosis' classes, as well as weighted and macro F1 scores. It also improved the nontumor class, significantly outperforming color jittering, Vahadane augmentation, and AIDA. Combining LDM-based augmentation with Macenko augmentation also improved model performance on the held-out test set. The LDM+Macenko model significantly outperformed the Macenko-only model, achieving higher weighted and macro F1 scores ($p_\text{raw}<0.01$, $p_\text{adjusted}<0.01$). Notably, across all models, differences in model performance in TCGA on the `good prognosis` class were not statistically significant between any pair of models.

\begin{table*}[htb]
\centering
\renewcommand{\arraystretch}{1.13}
\begin{tabular}{lcccccc}
\toprule
 & Good prognosis & Intermediate prognosis & Poor prognosis & Nontumor & Weighted & Macro \\
\midrule
UNI + logistic regression
  & 0.597 $\pm$ 0.080 & 0.660 $\pm$ 0.030 & 0.500 $\pm$ 0.077 & 0.739 $\pm$ 0.013 & 0.622 $\pm$ 0.029 & 0.624 $\pm$ 0.032 \\
No augmentation
  & 0.754 $\pm$ 0.028 & 0.655 $\pm$ 0.021 & 0.359 $\pm$ 0.113 & 0.797 $\pm$ 0.021 & 0.611 $\pm$ 0.046 & 0.641 $\pm$ 0.039 \\
Geometric transformation
  & 0.730 $\pm$ 0.064 & 0.639 $\pm$ 0.023 & 0.411 $\pm$ 0.060 & 0.791 $\pm$ 0.039 & 0.617 $\pm$ 0.014 & 0.643 $\pm$ 0.016 \\
Color jittering
  & 0.753 $\pm$ 0.043 & 0.632 $\pm$ 0.010 & 0.293 $\pm$ 0.091 & 0.767 $\pm$ 0.030 & 0.576 $\pm$ 0.022 & 0.611 $\pm$ 0.014 \\
Standard augmentation
  & 0.756 $\pm$ 0.030 & 0.639 $\pm$ 0.029 & 0.303 $\pm$ 0.134 & 0.772 $\pm$ 0.061 & 0.583 $\pm$ 0.061 & 0.617 $\pm$ 0.052 \\
Macenko stain augmentation
  & \underline{0.760 $\pm$ 0.030} & 0.654 $\pm$ 0.012 & 0.389 $\pm$ 0.118 & 0.801 $\pm$ 0.033 & 0.622 $\pm$ 0.032 & 0.651 $\pm$ 0.024 \\
Vahadane stain augmentation
  & 0.726 $\pm$ 0.032 & 0.672 $\pm$ 0.027 & 0.486 $\pm$ 0.101 & 0.775 $\pm$ 0.027 & 0.647 $\pm$ 0.042 & 0.665 $\pm$ 0.035 \\
CycleGAN stain normalization
  & 0.746 $\pm$ 0.020 & 0.669 $\pm$ 0.017 & 0.485 $\pm$ 0.037 & 0.802 $\pm$ 0.033 & 0.654 $\pm$ 0.016 & 0.675 $\pm$ 0.013 \\
AIDA feature alignment
  & 0.710 $\pm$ 0.022 & 0.646 $\pm$ 0.014 & 0.494 $\pm$ 0.054 & 0.755 $\pm$ 0.022 & 0.634 $\pm$ 0.024 & 0.651 $\pm$ 0.023 \\
\midrule
\makecell[l]{LDM synthetic augmentation \\\textbf{(proposed)}}
  & \textbf{0.769 $\pm$ 0.028} & 0.675 $\pm$ 0.017 & 0.526 $\pm$ 0.072 & \textbf{0.835 $\pm$ 0.019} & 0.680 $\pm$ 0.019 & \underline{0.701 $\pm$ 0.012} \\
\makecell[l]{LDM synthetic + Macenko \\ stain augmentation \textbf{(proposed)}}
  & 0.722 $\pm$ 0.027 & 0.683 $\pm$ 0.029 & 0.546 $\pm$ 0.047 & 0.833 $\pm$ 0.036 & \underline{0.683 $\pm$ 0.018} & 0.696 $\pm$ 0.018 \\
\makecell[l]{LDM synthetic + Vahadane \\stain augmentation \textbf{(proposed)}}
  & 0.731 $\pm$ 0.047 & \textbf{0.710 $\pm$ 0.021} & \textbf{0.592 $\pm$ 0.041} & 0.830 $\pm$ 0.020 & \textbf{0.706 $\pm$ 0.025} & \textbf{0.716 $\pm$ 0.027} \\
\bottomrule
\end{tabular}
\caption{Mean F1 score $\pm$ standard deviation for each class and averaged metrics (weighted and macro) across different augmentation strategies on the target cohort (TCGA).}
\label{tab:target_f1}
\end{table*}

\subsubsection{Cross-validation on the source cohort}

We also evaluated model performance on the source domain. Table~\ref{tab:source_f1} summarizes the mean and standard deviation of per-class and overall F1 scores on the cross-validation test sets from NLST. Among all models, the standard, LDM+Macenko, and LDM+Vahadane augmentations improved the macro F1 scores marginally, with no statistically significant difference. The LDM synthetic augmentation yielded slightly lower F1 scores; however, the differences were not significant in overall F1 scores or the per-class F1 scores ($p_\text{adjusted} > 0.05$). Overall, LDM-based augmentation improves generalization to the target cohort without sacrificing performance on the source cohort.

\begin{table*}[htb]
\centering
\renewcommand{\arraystretch}{1.13}
\begin{tabular}{lcccccc}
\toprule
 & Good prognosis & Intermediate prognosis & Poor prognosis & Nontumor & Weighted & Macro \\
\midrule
UNI + logistic regression 
  & 0.702 $\pm$ 0.098 & 0.664 $\pm$ 0.170 & 0.686 $\pm$ 0.153 & 0.662 $\pm$ 0.154 & 0.711 $\pm$ 0.119 & 0.678 $\pm$ 0.117 \\
No augmentation
  & 0.675 $\pm$ 0.134 & 0.721 $\pm$ 0.075 & 0.694 $\pm$ 0.125 & \underline{0.837 $\pm$ 0.071} & 0.763 $\pm$ 0.078 & 0.732 $\pm$ 0.062 \\
Geometric transformation
  & 0.694 $\pm$ 0.144 & 0.686 $\pm$ 0.084 & 0.611 $\pm$ 0.163 & 0.815 $\pm$ 0.057 & 0.714 $\pm$ 0.085 & 0.701 $\pm$ 0.073 \\
Color jittering
  & 0.678 $\pm$ 0.151 & 0.657 $\pm$ 0.078 & 0.604 $\pm$ 0.199 & 0.834 $\pm$ 0.049 & 0.718 $\pm$ 0.113 & 0.693 $\pm$ 0.093 \\
Standard augmentation
  & 0.694 $\pm$ 0.095 & 0.713 $\pm$ 0.103 & \underline{0.710 $\pm$ 0.138} & 0.812 $\pm$ 0.068 & \underline{0.770 $\pm$ 0.072} & 0.732 $\pm$ 0.056 \\
Macenko stain augmentation
  & \underline{0.718 $\pm$ 0.141} & \textbf{0.735 $\pm$ 0.072} & 0.640 $\pm$ 0.193 & 0.796 $\pm$ 0.070 & 0.758 $\pm$ 0.089 & 0.722 $\pm$ 0.071 \\
Vahadane stain augmentation
  & \textbf{0.754 $\pm$ 0.090} & 0.710 $\pm$ 0.080 & 0.688 $\pm$ 0.158 & 0.765 $\pm$ 0.121 & 0.761 $\pm$ 0.077 & 0.729 $\pm$ 0.069 \\
CycleGAN stain normalization
  & 0.700 $\pm$ 0.143 & 0.635 $\pm$ 0.134 & 0.631 $\pm$ 0.141 & 0.798 $\pm$ 0.078 & 0.717 $\pm$ 0.089 & 0.691 $\pm$ 0.077 \\
AIDA feature alignment
  & 0.702 $\pm$ 0.161 & 0.695 $\pm$ 0.071 & 0.608 $\pm$ 0.211 & 0.797 $\pm$ 0.074 & 0.737 $\pm$ 0.096 & 0.700 $\pm$ 0.088 \\
\midrule
\makecell[l]{LDM synthetic augmentation \\\textbf{(proposed)}}
  & 0.704 $\pm$ 0.141 & 0.649 $\pm$ 0.083 & 0.680 $\pm$ 0.161 & \textbf{0.842 $\pm$ 0.040} & 0.736 $\pm$ 0.111 & 0.719 $\pm$ 0.081 \\
\makecell[l]{LDM synthetic + Macenko \\stain augmentation \textbf{(proposed)}}
  & 0.714 $\pm$ 0.103 & \underline{0.724 $\pm$ 0.105} & \textbf{0.723 $\pm$ 0.144} & 0.805 $\pm$ 0.065 & \textbf{0.771 $\pm$ 0.089} & \textbf{0.741 $\pm$ 0.065} \\
\makecell[l]{LDM synthetic + Vahadane \\stain augmentation \textbf{(proposed)}}
  & 0.711 $\pm$ 0.083 & 0.716 $\pm$ 0.103 & 0.698 $\pm$ 0.184 & 0.811 $\pm$ 0.082 & 0.760 $\pm$ 0.096 & \underline{0.734 $\pm$ 0.075} \\
\bottomrule
\end{tabular}
\caption{Mean F1 score $\pm$ standard deviation for each class and averaged metrics (weighted and macro) across different augmentation strategies on the source cohort (NLST).}
\label{tab:source_f1}
\end{table*}

\subsection{Per-class domain shift analysis}

\begin{figure}[ht]
\centering
\includegraphics[width=0.43\textwidth]{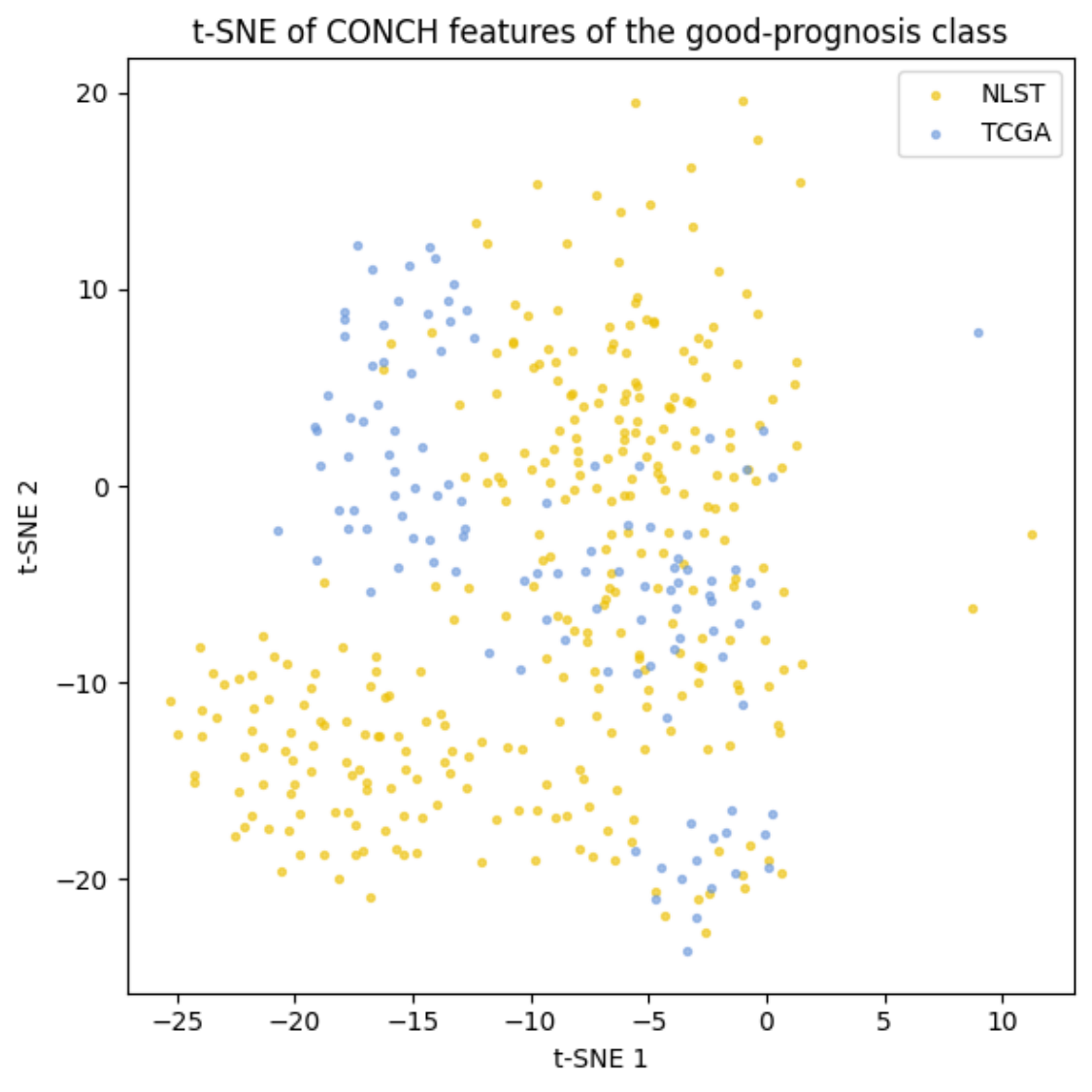}
\caption{
t-SNE visualization of CONCH feature embeddings for NLST and TCGA tiles in the `good prognosis' class. A substantial overlap exists between NLST and TCGA data points, indicating a smaller cross-cohort domain shift in this class.
}
\label{fig:good_prognosis}
\end{figure}

As shown in Sec.~\ref{sec:downstream_results}, all data augmentation methods had similar performance on the `good prognosis' class, which consists of tiles with the lepidic growth pattern. To investigate the class-level performance differences, we performed t-SNE analysis of CONCH~\cite{lu2024visual} feature embeddings of the `good prognosis' tiles from NLST and TCGA. As shown in Fig.~\ref{fig:good_prognosis}, the `good prognosis' tiles from the two cohorts showed substantial overlap, unlike the significant cross-cohort separation observed in Fig.~\ref{fig:t_sne}. This suggests that the class-level domain shift in the `good prognosis' class is less significant compared to the overall cohort-level domain shift, limiting the potential gain in domain generalization from data augmentation. This observation also aligns with the biological characteristics of the lepidic growth pattern. Lepidic tiles usually exhibit the growth of tumor cells along intact alveolar structures and have minimal stromal or pleural invasions, resulting in a highly distinctive and less varying morphology~\cite{moreira2020grading}. These features make the lepidic images less sensitive to cohort-specific variabilities, making it easier for DL models to learn truly meaningful patterns instead of cohort-specific noise.

\begin{table*}[t]
\centering
\renewcommand{\arraystretch}{1.13}
\begin{tabular}{lcccccc}
\toprule
 & Good prognosis & Intermediate prognosis & Poor prognosis & Nontumor & Weighted & Macro \\
\midrule
& \multicolumn{6}{c}{Cross-validation test set (NLST)} \\
\cmidrule(lr){2-7}
NLST
  & \textbf{0.743 $\pm$ 0.110} & 0.716 $\pm$ 0.073 & 0.662 $\pm$ 0.160 & 0.824 $\pm$ 0.038 & \textbf{0.761 $\pm$ 0.084} & \textbf{0.736 $\pm$ 0.065} \\
TCGA
  & 0.703 $\pm$ 0.149 & \textbf{0.720 $\pm$ 0.132} & \textbf{0.698 $\pm$ 0.181} & 0.787 $\pm$ 0.106 & 0.749 $\pm$ 0.140 & 0.727 $\pm$ 0.122 \\
NLST + TCGA
  & 0.704 $\pm$ 0.141 & 0.649 $\pm$ 0.083 & 0.680 $\pm$ 0.161 & \textbf{0.842 $\pm$ 0.040} & 0.736 $\pm$ 0.111 & 0.719 $\pm$ 0.081 \\
\midrule
& \multicolumn{6}{c}{Held-out test set from the target cohort (TCGA)} \\
\cmidrule(lr){2-7}
NLST
  & 0.719 $\pm$ 0.034 & 0.657 $\pm$ 0.016 & 0.364 $\pm$ 0.077 & 0.818 $\pm$ 0.022 & 0.614 $\pm$ 0.030 & 0.640 $\pm$ 0.028 \\
TCGA
  & 0.736 $\pm$ 0.033 & 0.668 $\pm$ 0.009 & 0.500 $\pm$ 0.070 & 0.822 $\pm$ 0.022 & 0.662 $\pm$ 0.021 & 0.681 $\pm$ 0.016 \\
NLST + TCGA
  & \textbf{0.769 $\pm$ 0.028} & \textbf{0.675 $\pm$ 0.017} & \textbf{0.526 $\pm$ 0.072} & \textbf{0.835 $\pm$ 0.019} & \textbf{0.680 $\pm$ 0.019} & \textbf{0.701 $\pm$ 0.012} \\
\bottomrule
\end{tabular}
\caption{Per-class and overall (weighted and macro) F1 scores (mean $\pm$ standard deviation) when the LDM is trained on unlabeled NLST only, unlabeled TCGA only, and unlabeled NLST+TCGA.}
\label{tab:ablation}
\end{table*}

\subsection{Ablation studies}

In our ablation studies, we compared three variants of the LDM to examine the effect of using unlabeled data from the target cohort to improve the generalizability of the downstream model. The three LDMs we compared are: (1) an LDM trained only on unlabeled data from NLST, (2) an LDM trained only on unlabeled data from TCGA, and (3) an LDM trained on unlabeled data from both NLST and TCGA. All other components in the pipeline remain unchanged except for the training data of the LDM. Table~\ref{tab:ablation} summarizes model performance on the NLST cross-validation test sets and the TCGA held-out test set. 

On the NLST cross-validation test sets, the NLST-only LDM achieved the highest source-cohort performance (weighted F1: 0.761).  No statistically significant differences in weighted F1 were observed among the three training conditions ($p = 0.09$--$0.71$). On the TCGA held-out test set, the LDM trained on NLST+TCGA achieved the best target-cohort performance (weighted F1: 0.680), significantly outperforming both the NLST-only LDM (weighted F1: 0.614, $p = 0.025$) and the TCGA-only LDM (weighted F1: 0.662, $p = 0.018$).

\section{Discussion}

DL models in computational pathology often struggle to generalize across different domains. Existing domain adaptation approaches either fail to leverage unlabeled data from the target domain or employ image-to-image translation methods, which can distort tissue structure and morphology, thereby degrading model performance. In this work, we present a semi-supervised domain adaptation framework based on latent diffusion models, which uses unlabeled images from both source and target domains to generate synthetic images in a target-aware manner. Our results show that the LDM-generated synthetic images, by bridging the gap between the two domains, improve the performance of downstream classifiers on the held-out test set from the target cohort without sacrificing performance on the source cohort. Models using synthetic images achieved substantially higher F1 scores compared to baseline augmentation methods.

The LDM-based data augmentation is effective in improving the generalizability of downstream models because it incorporates target-domain appearance while preserving realistic tissue morphology. By training the LDM on unlabeled data from both source and target domains, the LDM learns a generative prior that incorporates the difference between the two domains, including staining, scanners, or site-specific variations. In our conditioning mechanism, by conditioning the LDM on UNI features, we preserved tissue morphology. As a result, when using the LDM to generate target-aware synthetic images, the synthetic images retain the tissue morphology in the source images while exhibiting some characteristics of the target domain. Moreover, by conditioning the LDM on UNI features, we successfully generated synthetic images labeled by LUAD growth patterns using an LDM trained on unlabeled data. The label preservation assumption was empirically validated. As described in Section~\ref{sec:label_preservation}, the synthetic images achieved a Cohen's kappa of $0.744 \pm 0.062$ between the assigned labels and labels from an independent classifier, indicating a high agreement comparable to human inter-reader agreement. Moreover, we demonstrated the fidelity of the synthetic images with FID and showed their effectiveness in bridging the domain gap through a t-SNE analysis. Finally, using synthetic images, we exposed the downstream classifier to morphologically accurate yet appearance-augmented examples during training, thereby improving its generalization to the target cohort. 

Additionally, we found that stain augmentation complements LDM-based augmentation. This observation is consistent with how the two augmentation approaches create variabilities. First, although the LDM learned the staining distributions in NLST and TCGA, the synthetic images remain closely tied to the source morphology. This limits the degree of staining variation produced by the LDM: the changes are realistic and target-aware, but not too aggressive to make the resulting image unrealistic. In contrast, stain augmentation applies explicit perturbations to the hematoxylin and eosin stain vectors, enabling a broader range of color and staining variability than the LDM alone. Thus, stain augmentation complements the LDM by expanding variability in staining that the diffusion model produces. Second, the LDM-generated images have slight structural variations compared to the source images. They are morphologically consistent with the original tiles but are not pixel-wise replicas. In contrast, stain augmentation applies perturbations solely to the color and staining of the tiles, without altering the tissue structure. Together, these two augmentation types offer distinct types of variability: LDM provides target-aware appearance and subtle structural variation, while Vahadane offers strong, explicit staining variation. By combining these two, the downstream model is exposed to variabilities in staining, appearance, and tissue structure, resulting in a more effective data augmentation approach.  

In our per-class evaluation of downstream models, we found that all augmentation methods generally yielded no significant improvement in the `good prognosis' class and resulted in higher standard deviations in the `poor prognosis' class. For the `good prognosis' class, which consists of tiles with the lepidic growth pattern, our analysis suggests that the domain shift between NLST and TCGA for this class is smaller compared to the overall domain shift. This finding is consistent with the biological characteristics of the lepidic growth pattern, which refers to tumor cells growing along the alveoli without significant stromal or pleural invasion. This characteristic results in relatively lower variability in image appearance. As a result, data augmentation approaches provide limited additional benefits for this class compared to more heterogeneous classes, such as the `intermediate prognosis' and `poor prognosis' classes. For the `poor prognosis' class, across all models, we observed a higher standard deviation compared to other classes. This class consists of tiles with the solid and micropapillary growth patterns. Since tiles may contain tumors exhibiting both patterns, the tiles are more diverse in their tissue architecture. Additionally, these regions are more often affected by necrosis, inflammation, and desmoplasia compared to the lepidic pattern~\cite{moreira2020grading}, making it more difficult for DL models to learn consistent features and therefore resulting in higher standard deviations across the five folds.

The ablation studies highlight the distinct contributions of source- and target-domain unlabeled data to LDM training for effective SSDA. The suboptimal downstream performance of the NLST-only LDM on the TCGA held-out test set demonstrates that incorporating unlabeled target-domain data during LDM training is necessary for target-aware synthesis. In the TCGA-only experiment, the LDM was trained exclusively on target-domain images, yet during synthesis, it was conditioned on UNI embeddings extracted from source-domain (NLST) images. The fact that this did not hurt source-cohort performance suggests that an LDM trained on one cohort can generate plausible synthetic counterparts of images from an unseen cohort, which is possible because UNI embeddings encode morphological features that are shared across cohorts. Finally, the NLST+TCGA condition achieved the best target-cohort performance, demonstrating that training the LDM on unlabeled data from both domains is necessary for optimal target-cohort generalization and that source-domain data contributes meaningfully beyond what target-domain data alone provides.

Our findings have several implications. First, the proposed approach effectively enhances model generalization to external cohorts, as it improves cross-domain generalization without compromising performance on the source domain. Second, the proposed approach does not use labels from the target domain, making it suitable for scenarios such as pathology images, where unlabeled data are often available as annotations are costly or cannot be shared due to privacy concerns. 

There are several limitations of our work. First, the evaluation is limited to the NLST $\rightarrow$ TCGA direction, i.e., using NLST as the source cohort and TCGA as the target. We made this choice because of the limited sample size and class imbalance of the TCGA labeled set, which makes patient-level cross-validation splits unstable. Future work should expand the TCGA labeled set and test the reverse direction. Another limitation of this work is the computational cost. Diffusion models are computationally expensive, both during training and synthesis. Even though we attempted to alleviate this issue by using a latent-space diffusion model instead of an image-space model, the LDM remains computationally costly. In our experiments, training was performed on 2 NVIDIA L40S GPUs, each with 46GB of virtual memory, and required approximately 76 hours to complete. Moreover, generating a single synthetic image requires multiple passes through the U-Net denoiser, making it time-consuming to generate large-scale synthetic datasets. These computational requirements limit the usability of diffusion models in environments with limited computational resources. Additionally, while our analysis has shown that LDM-based data augmentation improves generalization, one needs to balance the tradeoff between better generalization and computation. In scenarios where access to the computational resources required for diffusion models is unavailable, stain augmentation, particularly Vahadane, can be a practical alternative.

\section{Conclusion}

In summary, this work demonstrates that utilizing latent diffusion models for SSDA facilitates improved generalization of pathology classification models to the target domain. By training an LDM on unlabeled images from both the source and target domains, the LDM can generate morphologically accurate and target-aware images. Using both target-aware synthetic images and traditional stain augmentation approaches, the downstream classifier is exposed to variations in tissue structure, image appearance, and staining, thereby achieving better generalization to the target domain. As part of our future work, we will explore conditioning mechanisms to better preserve details in tissue morphology and reduce noise in the labels of the synthetic data. Overall, this work highlights the utility of diffusion models in enhancing domain adaptation in computational pathology.

\section*{Conflict of interest statement}
WH reports receiving research funding from EarlyDiagnostics Inc and consulting fees from the Radiological Society of North America for editorial services. The remaining authors have no conflicts of interest to declare.


\section*{References}
\bibliographystyle{ieeetr} 
\bibliography{references}

@article{kingma2014adam,
  title={Adam: A method for stochastic optimization},
  author={Kingma, Diederik P},
  journal={arXiv preprint arXiv:1412.6980},
  year={2014}
}

@inproceedings{rombach2022high,
  title={High-resolution image synthesis with latent diffusion models},
  author={Rombach, Robin and Blattmann, Andreas and Lorenz, Dominik and Esser, Patrick and Ommer, Bj{\"o}rn},
  booktitle={Proceedings of the IEEE/CVF conference on computer vision and pattern recognition},
  pages={10684--10695},
  year={2022}
}

@inproceedings{
alfasly2025semantic,
title={Semantic and Visual Crop-Guided Diffusion Models for Heterogeneous Tissue Synthesis in Histopathology},
author={Saghir Alfasly and Wataru Uegami and MD ENAMUL HOQ and Ghazal Alabtah and Hamid Tizhoosh},
booktitle={The Thirty-ninth Annual Conference on Neural Information Processing Systems},
year={2025},
url={https://openreview.net/forum?id=yNVDkAjGjw}
}

@article{pezoulas2024synthetic,
  title={Synthetic data generation methods in healthcare: A review on open-source tools and methods},
  author={Pezoulas, Vasileios C and Zaridis, Dimitrios I and Mylona, Eugenia and Androutsos, Christos and Apostolidis, Kosmas and Tachos, Nikolaos S and Fotiadis, Dimitrios I},
  journal={Computational and structural biotechnology journal},
  volume={23},
  pages={2892--2910},
  year={2024},
  publisher={Elsevier}
}

@article{van2021deep,
  title={Deep learning in histopathology: the path to the clinic},
  author={Van der Laak, Jeroen and Litjens, Geert and Ciompi, Francesco},
  journal={Nature medicine},
  volume={27},
  number={5},
  pages={775--784},
  year={2021},
  publisher={Nature Publishing Group US New York}
}

@article{xu2019gan,
  title={GAN-based virtual re-staining: a promising solution for whole slide image analysis},
  author={Xu, Zhaoyang and Huang, Xingru and Moro, Carlos Fern{\'a}ndez and Boz{\'o}ky, B{\'e}la and Zhang, Qianni},
  journal={arXiv preprint arXiv:1901.04059},
  year={2019}
}

@article{kang2021stainnet,
  title={Stainnet: a fast and robust stain normalization network},
  author={Kang, Hongtao and Luo, Die and Feng, Weihua and Zeng, Shaoqun and Quan, Tingwei and Hu, Junbo and Liu, Xiuli},
  journal={Frontiers in Medicine},
  volume={8},
  pages={746307},
  year={2021},
  publisher={Frontiers Media SA}
}

@article{jiao2022staining,
  title={Staining condition visualization in digital histopathological whole-slide images},
  author={Jiao, Yiping and Li, Junhong and Fei, Shumin},
  journal={Multimedia Tools and Applications},
  volume={81},
  number={13},
  pages={17831--17847},
  year={2022},
  publisher={Springer}
}

@article{heusel2017gans,
  title={Gans trained by a two time-scale update rule converge to a local nash equilibrium},
  author={Heusel, Martin and Ramsauer, Hubert and Unterthiner, Thomas and Nessler, Bernhard and Hochreiter, Sepp},
  journal={Advances in neural information processing systems},
  volume={30},
  year={2017}
}

@article{ding2023tailoring,
  title={Tailoring pretext tasks to improve self-supervised learning in histopathologic subtype classification of lung adenocarcinomas},
  author={Ding, Ruiwen and Yadav, Anil and Rodriguez, Erika and Da Silva, Ana Cristina Araujo Lemos and Hsu, William},
  journal={Computers in biology and medicine},
  volume={166},
  pages={107484},
  year={2023},
  publisher={Elsevier}
}

@article{al2025cellomaps,
  title={CellOMaps: A compact representation for robust classification of lung adenocarcinoma growth patterns},
  author={Al-Rubaian, Arwa and Gunesli, Gozde N and Althakfi, Wajd A and Azam, Ayesha and Snead, David and Rajpoot, Nasir M and Raza, Shan E Ahmed},
  journal={Computers in Biology and Medicine},
  volume={192},
  pages={110127},
  year={2025},
  publisher={Elsevier}
}

@article{moreira2020grading,
  title={A grading system for invasive pulmonary adenocarcinoma: a proposal from the International Association for the Study of Lung Cancer Pathology Committee},
  author={Moreira, Andre L and Ocampo, Paolo SS and Xia, Yuhe and Zhong, Hua and Russell, Prudence A and Minami, Yuko and Cooper, Wendy A and Yoshida, Akihiko and Bubendorf, Lukas and Papotti, Mauro and others},
  journal={Journal of Thoracic Oncology},
  volume={15},
  number={10},
  pages={1599--1610},
  year={2020},
  publisher={Elsevier}
}

@article{tcga,
  title={The cancer genome atlas pan-cancer analysis project},
  author={Cancer Genome Atlas Research Network, JN and others},
  journal={Nat. Genet},
  volume={45},
  number={10},
  pages={1113--1120},
  year={2013}
}

@article{nlst,
  title={Reduced lung-cancer mortality with low-dose computed tomographic screening},
  author={National Lung Screening Trial Research Team},
  journal={New England Journal of Medicine},
  volume={365},
  number={5},
  pages={395--409},
  year={2011},
  publisher={Mass Medical Soc}
}

@article{russell2011does,
  title={Does lung adenocarcinoma subtype predict patient survival?: A clinicopathologic study based on the new International Association for the Study of Lung Cancer/American Thoracic Society/European Respiratory Society international multidisciplinary lung adenocarcinoma classification},
  author={Russell, Prudence A and Wainer, Zoe and Wright, Gavin M and Daniels, Marissa and Conron, Matthew and Williams, Richard A},
  journal={Journal of Thoracic Oncology},
  volume={6},
  number={9},
  pages={1496--1504},
  year={2011},
  publisher={Elsevier}
}

@inproceedings{graikos2024learned,
  title={Learned representation-guided diffusion models for large-image generation},
  author={Graikos, Alexandros and Yellapragada, Srikar and Le, Minh-Quan and Kapse, Saarthak and Prasanna, Prateek and Saltz, Joel and Samaras, Dimitris},
  booktitle={Proceedings of the IEEE/CVF Conference on Computer Vision and Pattern Recognition},
  pages={8532--8542},
  year={2024}
}

@inproceedings{yellapragada2025zoomldm,
  title={ZoomLDM: Latent Diffusion Model for multi-scale image generation},
  author={Yellapragada, Srikar and Graikos, Alexandros and Triaridis, Kostas and Prasanna, Prateek and Gupta, Rajarsi and Saltz, Joel and Samaras, Dimitris},
  booktitle={Proceedings of the Computer Vision and Pattern Recognition Conference},
  pages={23453--23463},
  year={2025}
}

@inproceedings{fang2023domain,
  title={A domain-invariant feature learning framework for histopathology images},
  author={Fang, Kun and Ding, Guangtai},
  booktitle={Third International Conference on Advanced Algorithms and Neural Networks (AANN 2023)},
  volume={12791},
  pages={392--397},
  year={2023},
  organization={SPIE}
}

@inproceedings{marini2021h,
  title={H\&E-adversarial network: a convolutional neural network to learn stain-invariant features through Hematoxylin \& Eosin regression},
  author={Marini, Niccolo and Atzori, Manfredo and Ot{\'a}lora, Sebastian and Marchand-Maillet, Stephane and M{\"u}ller, Henning},
  booktitle={Proceedings of the IEEE/CVF International Conference on Computer Vision},
  pages={601--610},
  year={2021}
}

@article{lafarge2019learning,
  title={Learning domain-invariant representations of histological images},
  author={Lafarge, Maxime W and Pluim, Josien PW and Eppenhof, Koen AJ and Veta, Mitko},
  journal={Frontiers in medicine},
  volume={6},
  pages={162},
  year={2019},
  publisher={Frontiers Media SA}
}

@article{otalora2019staining,
  title={Staining invariant features for improving generalization of deep convolutional neural networks in computational pathology},
  author={Ot{\'a}lora, Sebastian and Atzori, Manfredo and Andrearczyk, Vincent and Khan, Amjad and M{\"u}ller, Henning},
  journal={Frontiers in bioengineering and biotechnology},
  volume={7},
  pages={198},
  year={2019},
  publisher={Frontiers Media SA}
}

@article{du2025deeply,
  title={Deeply supervised two stage generative adversarial network for stain normalization},
  author={Du, Zhe and Zhang, Pujing and Huang, Xiaodong and Hu, Zhigang and Yang, Gege and Xi, Mengyang and Liu, Dechun},
  journal={Scientific Reports},
  volume={15},
  number={1},
  pages={7068},
  year={2025},
  publisher={Nature Publishing Group UK London}
}

@article{jose2021generative,
  title={Generative adversarial networks in digital pathology and histopathological image processing: a review},
  author={Jose, Laya and Liu, Sidong and Russo, Carlo and Nadort, Annemarie and Di Ieva, Antonio},
  journal={Journal of Pathology Informatics},
  volume={12},
  number={1},
  pages={43},
  year={2021},
  publisher={Elsevier}
}

@article{cong2022colour,
  title={Colour adaptive generative networks for stain normalisation of histopathology images},
  author={Cong, Cong and Liu, Sidong and Di Ieva, Antonio and Pagnucco, Maurice and Berkovsky, Shlomo and Song, Yang},
  journal={Medical Image Analysis},
  volume={82},
  pages={102580},
  year={2022},
  publisher={Elsevier}
}

@article{jahanifar2025domain,
  title={Domain generalization in computational pathology: survey and guidelines},
  author={Jahanifar, Mostafa and Raza, Manahil and Xu, Kesi and Vuong, Trinh Thi Le and Jewsbury, Robert and Shephard, Adam and Zamanitajeddin, Neda and Kwak, Jin Tae and Raza, Shan E Ahmed and Minhas, Fayyaz and others},
  journal={ACM Computing Surveys},
  volume={57},
  number={11},
  pages={1--37},
  year={2025},
  publisher={ACM New York, NY}
}

@article{sahiner2023data,
  title={Data drift in medical machine learning: implications and potential remedies},
  author={Sahiner, Berkman and Chen, Weijie and Samala, Ravi K and Petrick, Nicholas},
  journal={The British Journal of Radiology},
  volume={96},
  number={1150},
  pages={20220878},
  year={2023},
  publisher={Oxford University Press}
}

@article{zhang2025standardizing,
  title={Accelerating Data Processing and Benchmarking of AI Models for Pathology},
  author={Zhang, Andrew and Jaume, Guillaume and Vaidya, Anurag and Ding, Tong and Mahmood, Faisal},
  journal={arXiv preprint arXiv:2502.06750},
  year={2025}
}

@article{vaidya2025molecular,
  title={Molecular-driven Foundation Model for Oncologic Pathology},
  author={Vaidya, Anurag and Zhang, Andrew and Jaume, Guillaume and Song, Andrew H and Ding, Tong and Wagner, Sophia J and Lu, Ming Y and Doucet, Paul and Robertson, Harry and Almagro-Perez, Cristina and others},
  journal={arXiv preprint arXiv:2501.16652},
  year={2025}
}

@article{dosovitskiy2020image,
  title={An image is worth 16x16 words: Transformers for image recognition at scale},
  author={Dosovitskiy, Alexey},
  journal={arXiv preprint arXiv:2010.11929},
  year={2020}
}

@article{wolf2019huggingface,
  title={Huggingface's transformers: State-of-the-art natural language processing},
  author={Wolf, Thomas and Debut, Lysandre and Sanh, Victor and Chaumond, Julien and Delangue, Clement and Moi, Anthony and Cistac, Pierric and Rault, Tim and Louf, R{\'e}mi and Funtowicz, Morgan and others},
  journal={arXiv preprint arXiv:1910.03771},
  year={2019}
}

@inproceedings{macenko2009method,
  title={A method for normalizing histology slides for quantitative analysis},
  author={Macenko, Marc and Niethammer, Marc and Marron, James S and Borland, David and Woosley, John T and Guan, Xiaojun and Schmitt, Charles and Thomas, Nancy E},
  booktitle={2009 IEEE international symposium on biomedical imaging: from nano to macro},
  pages={1107--1110},
  year={2009},
  organization={IEEE}
}

@article{vahadane2016structure,
  title={Structure-preserving color normalization and sparse stain separation for histological images},
  author={Vahadane, Abhishek and Peng, Tingying and Sethi, Amit and Albarqouni, Shadi and Wang, Lichao and Baust, Maximilian and Steiger, Katja and Schlitter, Anna Melissa and Esposito, Irene and Navab, Nassir},
  journal={IEEE transactions on medical imaging},
  volume={35},
  number={8},
  pages={1962--1971},
  year={2016},
  publisher={IEEE}
}

@inproceedings{tsai2024test,
  title={Test-time stain adaptation with diffusion models for histopathology image classification},
  author={Tsai, Cheng-Chang and Chen, Yuan-Chih and Lu, Chun-Shien},
  booktitle={European Conference on Computer Vision},
  pages={257--275},
  year={2024},
  organization={Springer}
}

@article{runz2021normalization,
  title={Normalization of HE-stained histological images using cycle consistent generative adversarial networks},
  author={Runz, Marlen and Rusche, Daniel and Schmidt, Stefan and Weihrauch, Martin R and Hesser, J{\"u}rgen and Weis, Cleo-Aron},
  journal={Diagnostic Pathology},
  volume={16},
  number={1},
  pages={71},
  year={2021},
  publisher={Springer}
}

@article{AIDA,
  title={Learning generalizable AI models for multi-center histopathology image classification},
  author={Asadi-Aghbolaghi, Maryam and Darbandsari, Amirali and Zhang, Allen and Contreras-Sanz, Alberto and Boschman, Jeffrey and Ahmadvand, Pouya and K{\"o}bel, Martin and Farnell, David and Huntsman, David G and Churg, Andrew and others},
  journal={NPJ Precision Oncology},
  volume={8},
  number={1},
  pages={151},
  year={2024},
  publisher={Nature Publishing Group UK London}
}

@article{benjamini1995controlling,
  title={Controlling the false discovery rate: a practical and powerful approach to multiple testing},
  author={Benjamini, Yoav and Hochberg, Yosef},
  journal={Journal of the Royal statistical society: series B (Methodological)},
  volume={57},
  number={1},
  pages={289--300},
  year={1995},
  publisher={Wiley Online Library}
}

@article{xiao2021simultaneously,
  title={Simultaneously improve transferability and discriminability for adversarial domain adaptation},
  author={Xiao, Ting and Fan, Cangning and Liu, Peng and Liu, Hongwei},
  journal={Entropy},
  volume={24},
  number={1},
  pages={44},
  year={2021},
  publisher={MDPI}
}

@inproceedings{drexlin2025medi,
  title={Medi: Metadata-guided diffusion models for mitigating biases in tumor classification},
  author={Drexlin, David Jacob and Dippel, Jonas and Hense, Julius and Preni{\ss}l, Niklas and Montavon, Gr{\'e}goire and Klauschen, Frederick and M{\"u}ller, Klaus-Robert},
  booktitle={International Conference on Medical Image Computing and Computer-Assisted Intervention},
  pages={379--388},
  year={2025},
  organization={Springer}
}

@article{chen2024towards,
  title={Towards a general-purpose foundation model for computational pathology},
  author={Chen, Richard J and Ding, Tong and Lu, Ming Y and Williamson, Drew FK and Jaume, Guillaume and Song, Andrew H and Chen, Bowen and Zhang, Andrew and Shao, Daniel and Shaban, Muhammad and others},
  journal={Nature medicine},
  volume={30},
  number={3},
  pages={850--862},
  year={2024},
  publisher={Nature Publishing Group US New York}
}

@article{lu2024visual,
  title={A visual-language foundation model for computational pathology},
  author={Lu, Ming Y and Chen, Bowen and Williamson, Drew FK and Chen, Richard J and Liang, Ivy and Ding, Tong and Jaume, Guillaume and Odintsov, Igor and Le, Long Phi and Gerber, Georg and others},
  journal={Nature medicine},
  volume={30},
  number={3},
  pages={863--874},
  year={2024},
  publisher={Nature Publishing Group US New York}
}

@inproceedings{deng2009imagenet,
  title={Imagenet: A large-scale hierarchical image database},
  author={Deng, Jia and Dong, Wei and Socher, Richard and Li, Li-Jia and Li, Kai and Fei-Fei, Li},
  booktitle={2009 IEEE conference on computer vision and pattern recognition},
  pages={248--255},
  year={2009},
  organization={Ieee}
}

@inproceedings{yellapragada2024pathldm,
  title={Pathldm: Text conditioned latent diffusion model for histopathology},
  author={Yellapragada, Srikar and Graikos, Alexandros and Prasanna, Prateek and Kurc, Tahsin and Saltz, Joel and Samaras, Dimitris},
  booktitle={Proceedings of the IEEE/CVF Winter Conference on Applications of Computer Vision},
  pages={5182--5191},
  year={2024}
}

@article{TELLEZ2019101544,
title = {Quantifying the effects of data augmentation and stain color normalization in convolutional neural networks for computational pathology},
journal = {Medical Image Analysis},
volume = {58},
pages = {101544},
year = {2019},
issn = {1361-8415},
doi = {https://doi.org/10.1016/j.media.2019.101544},
url = {https://www.sciencedirect.com/science/article/pii/S1361841519300799},
author = {David Tellez and Geert Litjens and Péter Bándi and Wouter Bulten and John-Melle Bokhorst and Francesco Ciompi and Jeroen {van der Laak}},
}

\end{document}